\documentclass{article}
\usepackage{spconf,amsmath,graphicx,hyperref}

\usepackage{cite}
\usepackage{amsmath,amssymb,amsfonts}
\usepackage{algorithmic}
\usepackage{graphicx}
\usepackage{subfigure}
\usepackage{booktabs}
\usepackage{textcomp}
\usepackage{xcolor}
\usepackage{colortbl}
\usepackage{makecell}
\usepackage{xspace}
\usepackage{caption}
\newcommand{\alias}{MGRAL\xspace}

\newcommand{\ie}{\textit{i.e.\xspace}}
\newcommand{\eg}{\textit{e.g.\xspace}}
\definecolor{bestcolor}{gray}{.92}
\newcommand{\bestcell}[1]{\cellcolor{bestcolor}{#1}}

\title{Performance-guided Reinforced Active Learning for Object Detection}
\name{Zhixuan Liang$^{\star \ddagger}$\thanks{$\ddagger$ This work is partially done during Zhixuan's internship at SenseTime.} \qquad Xingyu Zeng$^{\dagger}$ \qquad Rui Zhao$^{\dagger}$ \qquad Ping Luo$^{\star}$}
\address{$^{\star}$ The University of Hong Kong \qquad
      $^{\dagger}$SenseTime Research}
\begin{document}
%
\maketitle
\begin{abstract}
Active learning (AL) strategies aim to train high-performance models with minimal labeling efforts, only selecting the most informative instances for annotation. Current approaches to evaluating data informativeness predominantly focus on the data's distribution or intrinsic information content and do not directly correlate with downstream task performance, such as mean average precision (mAP) in object detection. Thus, we propose Performance-guided (\ie mAP-guided) Reinforced Active Learning for Object Detection (\alias), a novel approach that leverages the concept of expected model output changes as informativeness.
To address the combinatorial explosion challenge of batch sample selection and the non-differentiable correlation between model performance and selected batches, \alias skillfully employs a reinforcement learning-based sampling agent that optimizes selection using policy gradient with mAP improvement as reward. Moreover, to reduce the computational overhead of mAP estimation with unlabeled samples, \alias utilizes an unsupervised way with fast look-up tables, ensuring feasible deployment.
We evaluate \alias's active learning performance on detection tasks over PASCAL VOC and COCO benchmarks. Our approach demonstrates the highest AL curve with convincing visualizations, establishing a new paradigm in reinforcement learning-driven active object detection.
\end{abstract}
\begin{keywords}
Active learning, reinforcement learning, active object detection
\end{keywords}
\section{Introduction}
\label{sec:intro}

While recent years have seen significant progress in model architectures, attention is increasingly shifting towards more efficient data utilization strategies. Among them, Active Learning (AL) stands out by training high-performance models with minimal labeling, particularly when field data arrives continuously and annotation is costly. By strategically selecting and annotating the most informative samples, active learning substantially improves data efficiency

\begin{figure}[htb]
\centerline{\includegraphics[width=.99\columnwidth]{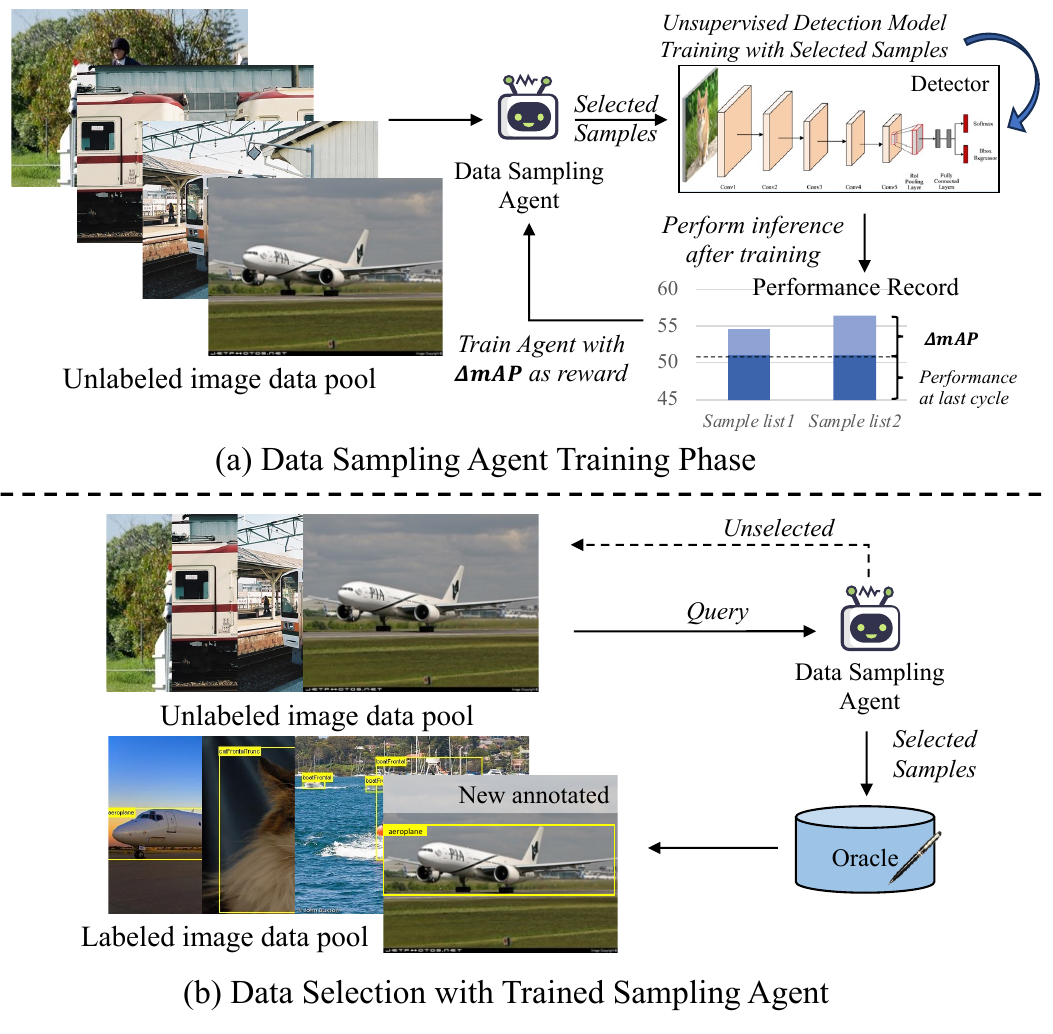}}
\vspace{-7pt}
\caption{\textbf{Overview of Performance-guided (mAP-guided) Reinforced Active Learning~(\alias) for object detection.} The RL-based agent learns to select most informative samples directly using performance gains ($\Delta \text{mAP}$) as reward.}
\vspace{-15pt}
\label{fig:pipeline}
\end{figure}

Therefore, the central question becomes how to identify most informative data. Prior methods select  the data points that are complementary to the currently labeled set in feature space, no matter characterized by distributional coverage or intrinsic ``information''. Although various definitions of this measure have been proposed and utilized as query strategies~\cite{LL4AL,CDAL,MIAL,zhang2021learning,EBAL,hua,mss,ppal,compas}, they do not align directly with downstream metrics, such as mean Average Precision (mAP) in object detection.
%
%
EMOC~\cite{freytag2014selecting} addresses this by measuring expected model output change, which matches our objective, but it was developed for Gaussian process regression and is not readily applicable to deep learning architectures.



In this paper, we propose Performance-guided (mAP-guided) Reinforced Active Learning (\alias) for object detection, directly using mAP changes to guide data sampling. However, different from EMOC not using deep learning downstream task model, our optimization pipeline must handle discrete, non-differentiable batch selection from the unlabeled pool, which makes it unable to update the selected samples through gradient from downstream metrics straightly. We therefore employ a reinforcement learning-based sampling agent that uses mAP variation ($\Delta$mAP) as the reward to enable optimizing the selection process through policy gradient technique. This approach achieves efficient exploration of possible batch combinations, maximizing mAP improvement per selected batch.

Because candidate samples are unlabeled when estimating potential mAP changes, we substitute an unsupervised surrogate for the detector to approximate downstream performance. It is effective as we only need to capture the relative influence among candidate data batches instead of correct labels to guide the active learning process.
To mitigate the high cost made by reinforcement learning (RL)-based agent of retraining the whole detector from scratch once per RL iteration, we introduce a fast lookup-table accelerator. Empirical results demonstrate our AL method's efficiency on both Pascal VOC~\cite{PASCALVOC} and MS COCO~\cite{COCO} across two detector backbones. Visualizations further show that selection criteria should vary across training stages, and only downstream performance provides a reliable signal.

Our contributions can be summarized as:
(1) We propose~\alias, extending the concept of expected model output changes for deep active learning, directly aligning the sampling strategy with mAP improvement.
(2) We introduce a reinforcement learning-based agent for efficient batch selection, addressing the combinatorial explosion and non-differentiable challenge in optimizing from mAP improvement.
(3) We develop practical techniques for~\alias, and evaluate it on multiple downstream benchmarks and different backbones, revealing our method's superior performance and the key effect that expected model output change played in it.


\section{Related Works}


\subsection{Active Learning for Object Detection}
Earliest attempts in active learning for object detection include LL4AL~\cite{LL4AL} adapting instance loss predictions and AL4DeepDetection~\cite{aghdam2019active} combining uncertainty metrics for foreground objects and background pixels. CDAL~\cite{CDAL} then enhances sample representativeness through spatial context, while MIAL~\cite{MIAL,wan2023multiple} employs adversarial classifiers and an unsupervised framework. EBAL~\cite{EBAL} integrates uncertainty and diversity but faces challenges with computational complexity and class imbalance. More recent works including ComPAS~\cite{compas}, MEH+HUA~\cite{hua}, MSS~\cite{mss} and PPAL~\cite{ppal} focus more on efficient sample uncertainty definition and provides new solutions to batch selection problem. 

However, these methods do not directly align with the task model's performance.
Our work, instead, directly utilizes mAP improvement to guide the selection process, addressing the limitations of previous approaches in balancing various metrics and handling batch selection effectively.

\subsection{Reinforcement Learning in Active Learning}

Reinforcement learning has been adopted to learn better query strategies in active learning, aiming to maximize task model performance. Techniques include imitation learning~\cite{liu2018learning}, bi-directional RNNs~\cite{contardo2017meta}, Deep Q-Networks~\cite{konyushkova2018discovering,DBLP:conf/iclr/CasanovaPRP20}, and policy gradient for simultaneous learning of data representation and selection heuristics~\cite{bachman2017learning} have been utilized to address data representativeness and develop efficient sampling strategies.
However, these methods struggle with integrating multi-instance uncertainty within a single image and batch sample selection for object detection tasks. Our approach addresses these challenges by leveraging $\Delta \text{mAP}$ as the reward to optimize the batch sample selection strategy.


\section{Methodology}

\subsection{\alias Active Learning Pipeline}
As depicted in Fig.~\ref{fig:pipeline}, to handle the non-differentiable link between sample selection and downstream performance changes, we embed a reinforcement learning-based sampling agent into classical pool-based active learning framework, optimizing selections for mAP improvement.

\textbf{Training.} \alias uses a nested optimization scheme: the \emph{outer} RL loop trains the sampler; the \emph{inner} detector loop estimates mAP gains for each RL step. For each AL round (cycle $t$), \textbf{[Inner Loop]} we build a lookup table by training multiple detector variants in parallel (Sec.~\ref{subsec:acceleration}). The results are pre-recorded and serve for efficiently approximating performance gains. \textbf{[Outer Loop]} The sampling agent (Sec.~\ref{subsec:agent_architect}) then runs RL iterations ($i$), selecting candidate sets from the unlabeled pool and receiving feedback from the estimated mAP improvements (Sec.~\ref{subsec:reward_design}). This feedback updates the policy via our stabilized policy-gradient optimization (Sec.~\ref{subsec:training}).

\textbf{Inference (active selection).} The trained agent traverses the unlabeled pool with its sequence-model-based architecture, computing for each image a selection score from image features and historical selection context, and finally chooses the top-$B$ (budget) images for annotation.

This design enables \alias to efficiently bridge sample selection and detector performance gains.

\subsection{\alias Data Sampling Agent}
\label{subsec:agent_architect}
\begin{figure}[t]
    \includegraphics[width=0.95\columnwidth]{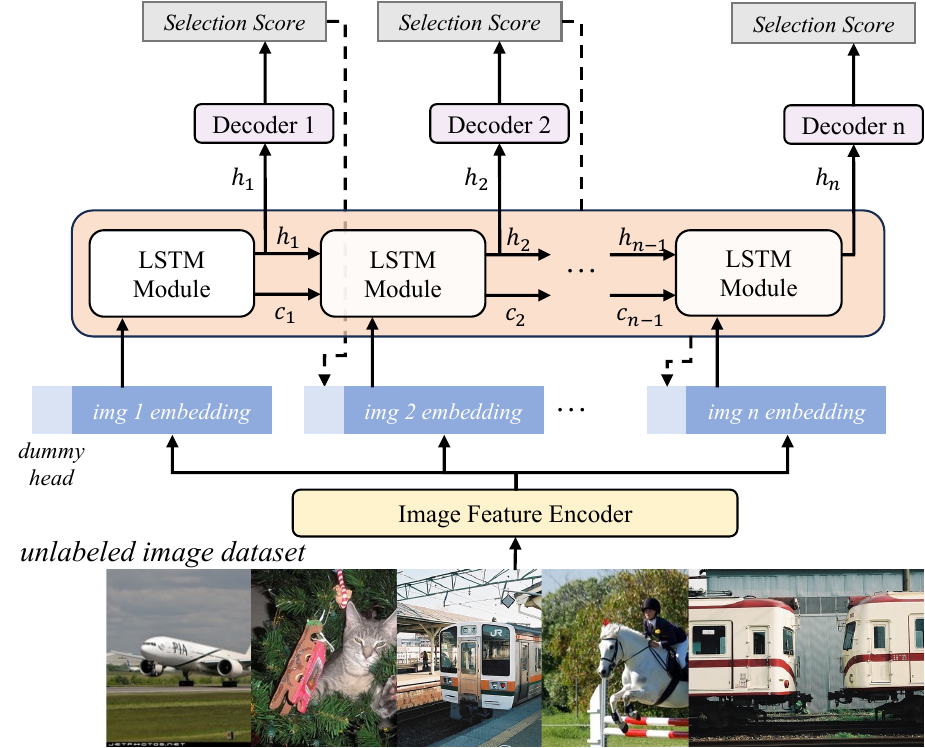}
    \vspace{-3pt}
    \caption{\textbf{Data sampling agent architecture.}}
	\label{fig:controller}
    \vspace{-10pt}
\end{figure}

The sampling agent adopts a Neural Architecture Search-inspired design, utilizing Long Short-Term Memory (LSTM)~\cite{hochreiter1997long} network for sequential sample selection. As the architecture of sampling agent shown in Fig.~\ref{fig:controller}, each image $I_k$ is first processed through a pre-trained detection encoder $\Phi(\cdot)$, which derives a feature vector $\Phi(I_k)$. This architecture considers both new data features and previous selection patterns through combining each image's embedding with the preceding unit's decision vector. And then, the extracted representation flows through parameter-sharing LSTM modules that ensure sequential modeling while preventing gradient vanishing. After LSTM, a 2-layer MLP decoder $\Psi_k(\cdot)$ (top in Fig.~\ref{fig:controller}) outputs selection scores from LSTM hidden states for each sample.

After processing the whole sequence, our algorithm selects the top-$B$ (budget) samples that achieves highest scores from the unlabeled pool. This selection mechanism forms the basis of our reward computation and policy updates.

\subsection{Performance-Driven Reward Design}
\label{subsec:reward_design}
The reward mechanism is the cornerstone of our approach, which links the sample selection directly with detector performance improvement ($\Delta \text{mAP}$). Detailedly, at active learning cycle $t$, for each RL training iteration $i$, \alias first selects one batch of sample candidates (without labels) and combines them with currently labeled dataset. Assuming we have labels of these selected candidates, we can train a new detector from scratch with the combined dataset to compute $\text{mAP}_i^t$. Then,
\begin{equation}
\vspace{-2pt}
    \Delta\text{mAP}_i^t = \text{mAP}_i^t - \text{mAP}^{t}_{i-1}.
    \label{eq:delta_map}
\end{equation}

However, candidates are only annotated after selection by the trained RL agent $\theta^t_{\text{agent}}$.
To address this, we employ a semi-supervised detection model trained on labeled $X_{L}^{t}$ and unlabeled $X_{S,i}^t$ to approximate the performance of a fully-supervised detector trained on $X_{L}^{t} + X_{S,i}^t$. In practice, various semi-supervised methods (\eg~ISD-SSD~\cite{jeong2021interpolation}) can be adopted depending on the detector backbone, following an objective: $L_{\text{Total}} = L_S + w(t) \times L_U$, where $L_S$ denotes supervised loss on labeled data, $L_U$ represents unsupervised consistency losses on unlabeled data, and $w(t)$ is a time-dependent weight that gradually increases during training. After that, we only need to replace ${\text{mAP}}_i^t$ in Eq.~\ref{eq:delta_map} with this approximated term.
This manner is sufficient as we only need to rank and identify the most potentially impactful samples for query strategy, not make exact performance predictions.


\subsection{Stabilized Policy-gradient Optimization}
\label{subsec:training}
As illustrated above, the training of \alias centers on iterative refinement of RL-based data sampling agent.
Using \textbf{policy gradient}, we enable model output changes to flow back to this discrete selection process, allowing effective update of the agent's parameters.
Moreover, to stabilize the policy gradient training, we employ a moving average baseline to normalize the reward signal and reduce variance in the policy gradient updates. Let $\text{mAP}^t_i$ denote the mAP achieved at AL cycle $t$ and RL iteration $i$ as above. We additionally maintain a reference $\text{mAP}_{ref}$ baseline to track the performance trend and smooth the $\Delta \text{mAP}_i^t$ signal as:
\begin{equation}
\vspace{-2pt}
\text{mAP}_{ref,i}^t = \lambda * \text{mAP}_{ref, i-1}^{t} + (1 - \lambda) * (\text{mAP}_i^t -\text{mAP}_{ref,i-1}^{t}),
\label{eq:map_ref}
\end{equation}
where $\lambda$ is the momentum hyper-parameter that controls the update rate.
Then, we substitute this reference $\text{mAP}^{t}_{ref,i-1}$ to the actual $\text{mAP}^{t}_{i-1}$ in Eq.~\ref{eq:delta_map} and rewrite policy gradient loss:
\begin{equation}
loss_{\theta_{\text{agent}},i} = -\Delta \text{mAP}_i^t = -(\text{mAP}_i^t -\text{mAP}_{ref,i-1}^{t}).
\end{equation}
This normalized reward indicates whether the current batch selection is better or worse than the recent average, enabling effective optimization over the discrete selection process.




\begin{figure*}[t] \centering
\begin{minipage}{0.58\textwidth}
\vspace{-15pt}
\subfigure[SSD on Pascal VOC] {
\label{fig:voc_result}   
\includegraphics[width=0.48\textwidth]{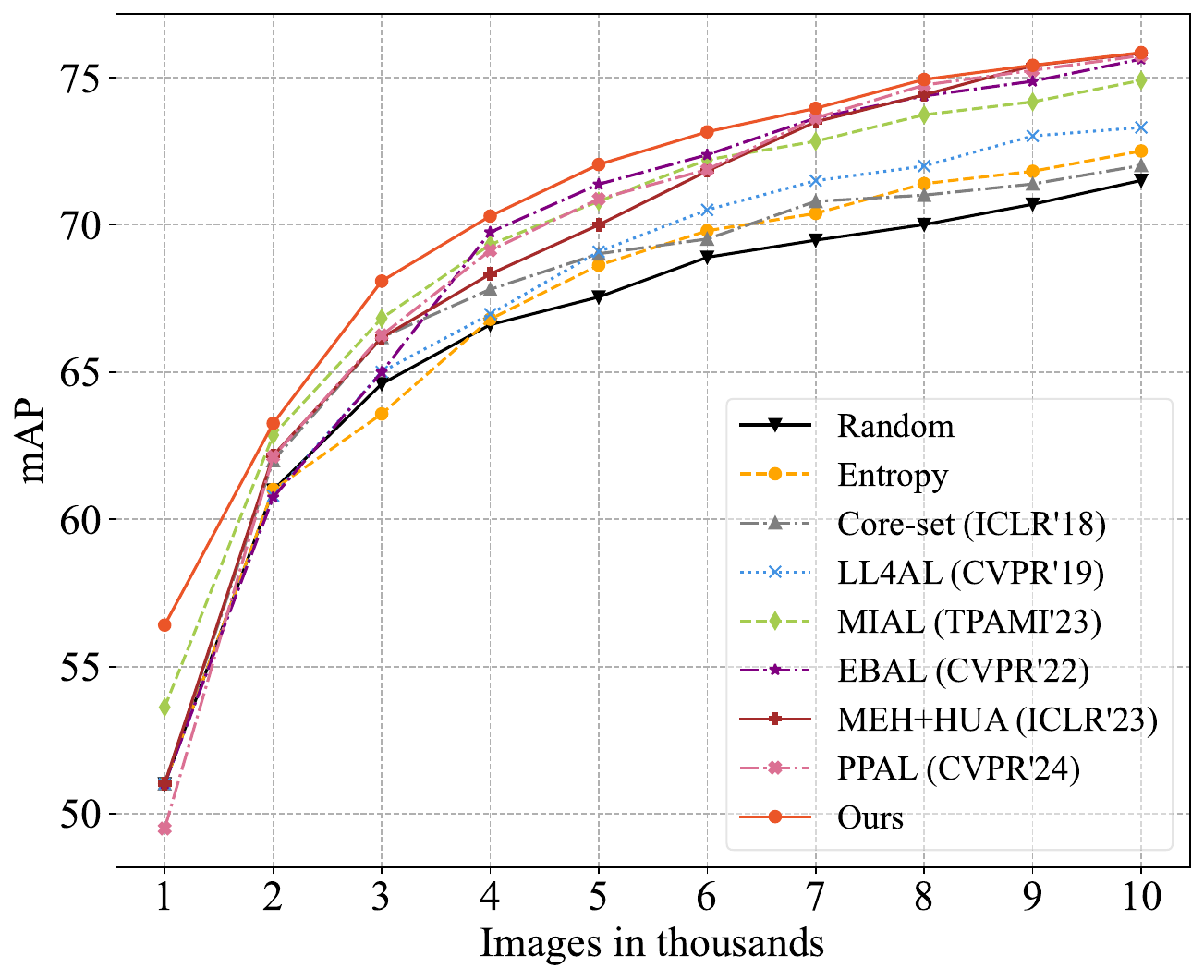}  
}
\hspace{-8pt}
\subfigure[RetinaNet on MS COCO] { 
\label{fig:coco_result} 
\includegraphics[width=0.48\textwidth]{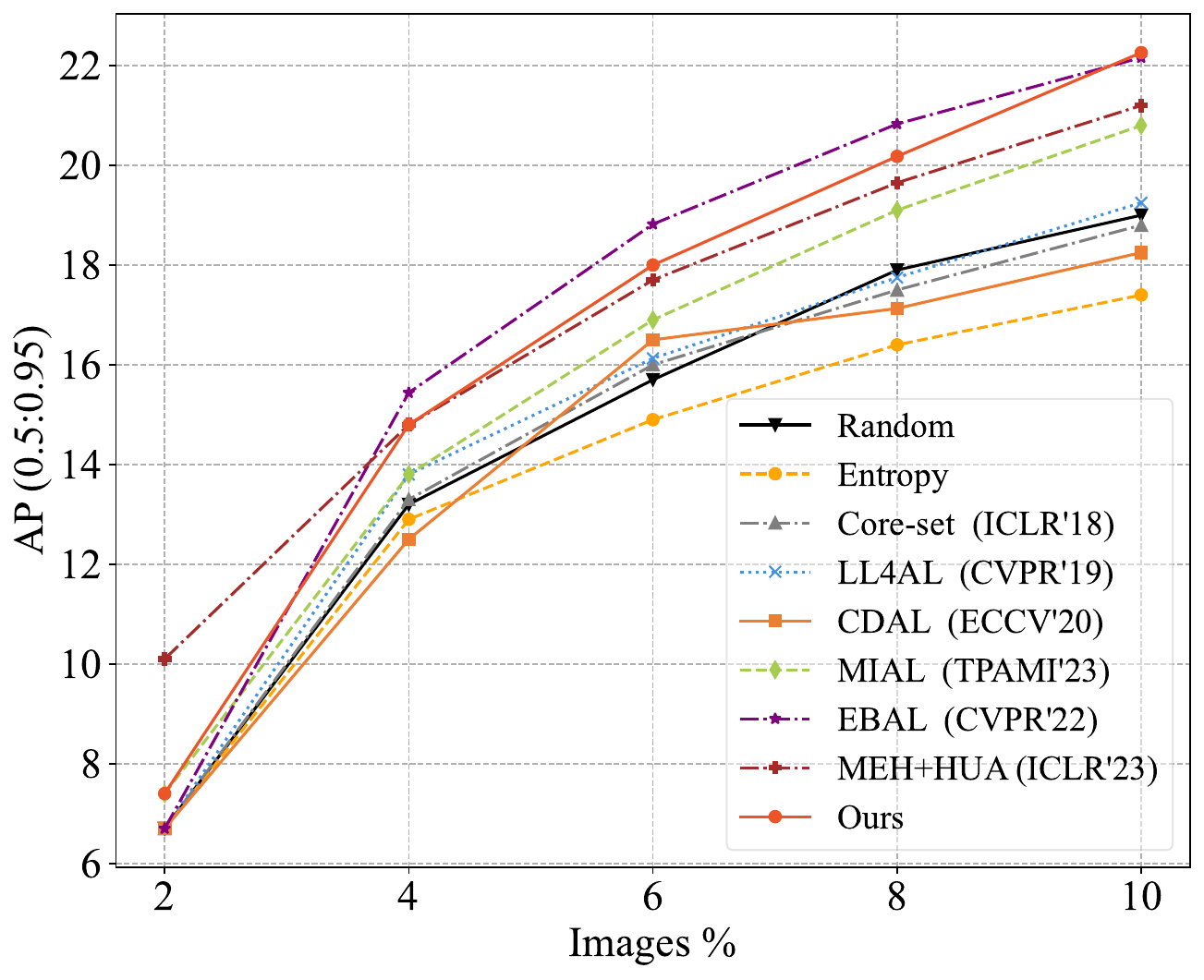}  
}
\vspace{-8pt}
\caption{\textbf{Comparative performance of active learning methods.} The mAP is plotted against the number of labeled images.} 
\label{fig:overall_result}
\vspace{-10pt}
\end{minipage}
\hfill
\begin{minipage}{0.38\textwidth}
\centering
  \small
  \captionof{table}{\textbf{Training time of \alias with and without lookup table on Pascal VOC.}}
  \label{tab:lookup_table_time}
  \vspace{-5pt}
  \resizebox{\textwidth}{!}{
  \begin{tabular}{c|c|c}
    \toprule
    \textbf{Method} & \textbf{Time of 10 Iters} & \textbf{Time for One Cycle}\\
    \midrule
    w/o acceleration & 4800 min & unknown\\
    \bestcell{\textbf{w/ lookup table}} & \bestcell{\textbf{3 min}} & \bestcell{\textbf{640 min}}\\
  \bottomrule
\end{tabular}}

  \label{tab:acceleration}
    \centering
    \small
  \captionof{table}{\textbf{Ablation on Efficiency analysis.} All results are about models tested on VOC.}
  \resizebox{\textwidth}{!}{
  \begin{tabular}{c|c|c|c}
    \toprule
    \textbf{Method} & \textbf{\makecell{Training Time\\of One Cycle}} & \textbf{\makecell{Inference Time\\of One Cycle}} & \textbf{\# of params}\\
    \midrule
    Entropy & 0 & 5 min & 26.3 M\\
    MIAL~\cite{MIAL} & 7 h 13 min & 43 min & 31.9 M\\
    EBAL~\cite{EBAL} & 0 & 181 min & 26.5 M\\
    \bestcell{\textbf{Ours}} & \bestcell{\textbf{(9 h) + 7 h 8 min}} & \bestcell{\textbf{0.5 min}} & \bestcell{\textbf{33.0 M}}\\
  \bottomrule
\end{tabular}}
  \label{tab:efficiency}
\end{minipage}
\vspace{-8pt}
\end{figure*}

\begin{figure}[t]
\centering
\includegraphics[width=1.03\columnwidth]{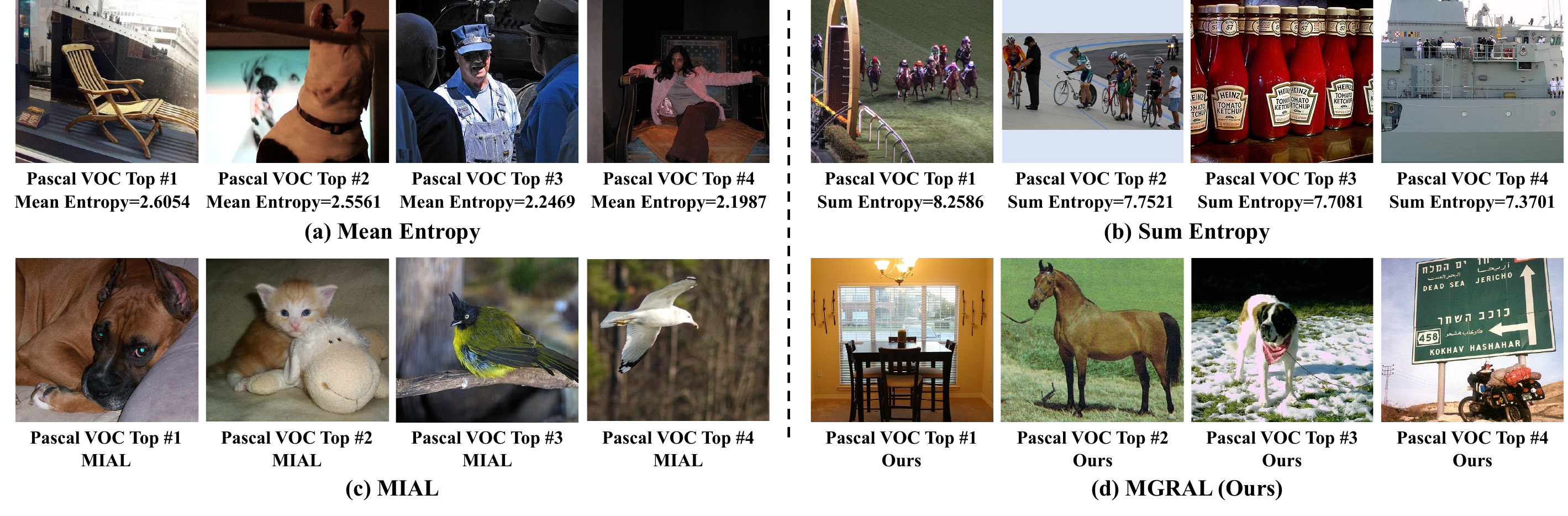}
\vspace{-22pt}
\caption{\textbf{Visualizations of the selected samples with highest scores by different AL methods during first cycle on VOC.}}
\vspace{-10pt}
\label{fig:visualization}
\end{figure}

\subsection{Acceleration Technique}
\label{subsec:acceleration}
Another critical challenge of our approach lies in the computational cost of our inner loop optimization. By default, \alias's each RL iteration requires a complete retraining of the unsupervised detector to estimate mAP for one candidate batch selection, yielding a prohibitive computation of 2200 (VOC) and 800 (COCO) sequential trainings per AL cycle.

We introduce a fast lookup-table (LUT) technique for acceleration. Before each cycle's agent training, we train $M$ unsupervised models \textit{in parallel} and record their performance, each on the current labeled data $X_{L}^t$ plus a randomly selected batch of unlabeled data $\tilde{X}_{S,l}^t$ ($l=1,2,...,M$) with the size $|\tilde{X}_{S,l}^t|$ equal to AL budget $B$. 
%
Then during the agent's RL training, instead of retraining the detector again and again, we compute the \textit{Wasserstein distance}~\cite{vallender1974calculation} (an optimal-transport metric) between visual representations of the agent-selected batch and LUT entries, and approximate mAP via a distance-weighted sum of the most similar records (weights inversely proportional to distance).
If no entry meets a preset threshold (\eg, mean minus one standard deviation), we fall back to direct retraining.
This acceleration converts thousands of sequential trainings into one parallel phase per cycle, delivering $>1000\times$ speedup and making \alias practical.  


\section{Experiments}
\subsection{Experiment Details}
Datasets and protocol: VOC07+12 and COCO2017. VOC-1k labeled init, +1k/cycle for 10 cycles; COCO-2.0\% init, +2.0\%/cycle to 10.0\%.
Detectors: SSD~\cite{SSD} on VOC; RetinaNet~\cite{lin2017focal} on COCO.
Agent: 257-dimension hidden (256 image embedding + 1 prev-score) with an LSTM of equal width; Adam (lr $3.5 \times 10^{-4}$). 
RL iters per cycle: 2,200 (VOC), 800 (COCO); baseline decay $\lambda=0.5$ for mAP$_{ref}$.
LUT: per cycle, $M$=200 ISD-SSD runs on VOC; on COCO, 30 SED-SSOD runs (150 records in total).

\vspace{-2pt}
\subsection{Overall Performance}
We compare \alias against random, entropy, Core-set~\cite{coreset}, CDAL~\cite{CDAL}, LL4AL~\cite{LL4AL}, MIAL~\cite{MIAL,wan2023multiple}, EBAL (\textit{aka} DivProto)~\cite{EBAL}, MEH+HUA~\cite{hua}, and PPAL~\cite{ppal}. Results in Fig.~\ref{fig:overall_result} show that on VOC, \alias consistently outperforms all methods, validating our mAP-guided design. On COCO, \alias surpasses most baselines. Despite a slight early lag vs.~EBAL, \alias exhibits the steepest growth and finishes ahead. We attribute this to EBAL's feature learning benefiting from more labels (weaker early, stronger mid-stage), whereas our performance-guided sampling yields larger early gains and better label-cost efficiency-especially clear on VOC. On COCO (117k vs.~VOC's 16.5k), the larger unlabeled pool induces initial parity; \alias then accelerates and takes the lead. This indicates \alias excels at fine-grained selection; on very large pools, a hierarchical scheme (\eg, clustering to prune the search space) could further amplify gains.
In summary, mAP-as-reward directly optimizes detection, and the RL formulation can easily generalize to other tasks by changing the reward metric beyond detection.


\vspace{-2pt}
\subsection{Visualization Analysis}
We visualize selections at the first AL cycle on PASCAL VOC (Fig.~\ref{fig:visualization}). It shows mean-entropy sampling favors exposed/blurred images (large entropy drop), which can hinder early training; sum-entropy sampling picks images with many instances but from few categories, causing redundancy. MIAL~\cite{MIAL} and \alias both prefer single, centered objects that provide clean supervision, while \alias further promotes category diversity, improving robustness across classes. These patterns explain \alias's higher mAP and validate its performance-guided, diversity-aware selection for building balanced early datasets.

\subsection{Ablation on Efficiency}
Our lookup table (LUT) accelerator slashes time cost: for 10 iterations on four GTX 1080Ti GPUs, training drops from 4,800 minutes to 3 minutes (Tab.~\ref{tab:lookup_table_time}). Although \alias incurs a slightly higher one-time setup than MIAL~\cite{MIAL} and EBAL~\cite{EBAL}, it is more efficient in subsequent cycles due to shorter inference (Tab.~\ref{tab:efficiency}), which aligns with offline AL workflows where collection/selection/labeling are batched rather than real time. Space-wise, \alias remains competitive ($33.0$M params vs.~EBAL $26.5$M, $31.9$M), and the LUT adds only $\sim12$KB while enabling the above acceleration.



\section{Conclusion}
We present \alias, a Performance-guided Reinforced Active Learning framework for object detection. By using $\Delta\text{mAP}$ as reward, our method effectively aligns batch selection with detection performance improvement, addressing the non-differentiable nature of selection process through policy gradient. The integration of unsupervised approximation and lookup table acceleration enables practical deployment while maintaining efficiency. \alias achieves consistent improvements over prior methods on both VOC and COCO across different backbones, establishing a new paradigm that combines performance-driven reinforcement learning and efficient active learning for object detection.
Exploring early stopping techniques for faster mAP estimation and replacing lookup tables with prediction networks for online reinforcement learning are promising directions for future work.

\clearpage
\bibliographystyle{IEEEbib}
\bibliography{strings,refs}

\end{document}